\documentclass[sigconf]{acmart}

\usepackage{booktabs} % For formal tables

\usepackage{url, graphicx, subfigure}
\usepackage{amsmath, amsthm, amssymb}
\usepackage{algorithm,algorithmic,paralist}
\usepackage{bm}
\usepackage{multirow}

\newtheorem{thm}{Theorem}

\newtheorem{lemma}{Lemma}

\newtheorem{definition}{Definition}
\def \R {\mathbb{R}}

\def \x {\mathbf{x}}
\def \w {\mathbf{w}}
\def \X {\mathbf{X}}

\def \Z {\mathbf{Z}}

\def \E {\mathbb{E}}

\def \M {\mathbf{M}}

\def \hX {\widehat{X}}

\def \L {\mathcal{L}}
\def \bL {\bar{\mathcal{L}}}

\def \bhX {\widehat{\mathbf{X}}}
\def \A {\mathbf{A}}
\def \B {\mathbf{B}}
\def \U {\mathbf{U}}
\def \V {\mathbf{V}}
\def \bSigma{\bm\Sigma}

\def \y {\mathbf{y}}

\newcommand{\argmin}{\operatornamewithlimits{argmin}}
\newcommand{\argmax}{\operatornamewithlimits{argmax}}

\begin{document}

\settopmatter{printacmref=false} % Removes citation information below abstract
\renewcommand\footnotetextcopyrightpermission[1]{} % removes footnote with conference information in first column
\pagestyle{plain} % removes running headers

\title{Active Feature Acquisition with Supervised Matrix Completion}
%\titlenote{Produces the permission block, and copyright information}

\author{Sheng-Jun Huang}
\affiliation{%
	\institution{Nanjing University of Aeronautics and Astronautics}
	\streetaddress{29 Jiangjun Road}
	\city{Nanjing}
	\state{China}
	\postcode{211106}
}
\email{huangsj@nuaa.edu.cn}

\author{Miao Xu}
\affiliation{%
	\institution{RIKEN Center for AIP}
	\streetaddress{103-0027}
	\city{Tokyo}
	\state{Japan}
	\postcode{}
}
\email{miao.xu@riken.jp}

\author{Ming-Kun Xie}
\affiliation{%
	\institution{Nanjing University of Aeronautics and Astronautics}
	\streetaddress{29 Jiangjun Road}
	\city{Nanjing}
	\state{China}
	\postcode{211106}
}
\email{mkxie@nuaa.edu.cn}

\author{Masashi Sugiyama}
\affiliation{%
	\institution{RIKEN Center for AIP\\University of Tokyo}
	\streetaddress{103-0027}
	\city{Tokyo}
	\state{Japan}
	\postcode{}
}
\email{sugi@k.u-tokyo.ac.jp}

\author{Gang Niu}
\affiliation{%
	\institution{RIKEN Center for AIP}
	\streetaddress{103-0027}
	\city{Tokyo}
	\state{Japan}
	\postcode{}
}
\email{gang.niu@riken.jp}

\author{Songcan Chen}
\affiliation{%
	\institution{Nanjing University of Aeronautics and Astronautics}
	\streetaddress{29 Jiangjun Road}
	\city{Nanjing}
	\state{China}
	\postcode{211106}
}
\email{s.chen@nuaa.edu.cn}

\renewcommand{\shortauthors}{S.-J. Huang et al.}

\fancyhead{}

\begin{abstract}
	Feature missing is a serious problem in many applications, which may lead to low quality of training data and further significantly degrade the learning performance. While feature acquisition usually involves special devices or complex processes, it is expensive to acquire all feature values for the whole dataset. On the other hand, features may be correlated with each other, and some values may be recovered from the others. It is thus important to decide which features are most informative for recovering the other features as well as improving the learning performance. In this paper, we try to train an effective classification model with the least acquisition cost by jointly performing \emph{active feature querying} and \emph{supervised matrix completion}. When completing the feature matrix, a novel objective function is proposed to simultaneously minimize the reconstruction error on observed entries and the supervised loss on training data. When querying the feature value, the most uncertain entry is actively selected based on the variance of previous iterations. In addition, a bi-objective optimization method is presented for \emph{cost-aware active selection} when features bear different acquisition costs. The effectiveness of the proposed approach is well validated by both theoretical analysis and experimental study.
\end{abstract}

%\begin{CCSXML}
%	<ccs2012>
%	<concept>
%	<concept_id>10010147.10010257.10010282.10011304</concept_id>
%	<concept_desc>Computing methodologies~Active learning settings</concept_desc>
%	<concept_significance>500</concept_significance>
%	</concept>
%	<concept>
%	<concept_id>10002951.10003227.10003351</concept_id>
%	<concept_desc>Information systems~Data mining</concept_desc>
%	<concept_significance>500</concept_significance>
%	</concept>
%	</ccs2012>
%\end{CCSXML}

%\ccsdesc[500]{Computing methodologies~Active learning settings}
%\ccsdesc[500]{Information systems~Data mining}

%\keywords{Active learning, feature acquisition, matrix completion}

\maketitle

\section{Introduction}
In data mining and machine learning tasks, a set of data objects is usually represented as a feature matrix, where each row is an object and each column is one dimension of the features. In many applications, the feature matrix may be partially observed with missing values due to various reasons \cite{liu1998feature}. For example, in disease diagnosis, a patient is an object, and the feature space consists of the physical examination results. Then some patients may selectively take some of the examinations, leaving the other features missing \cite{lim2005hybrid}. In wireless sensor network analysis, multiple sensors detect features of the environment in different aspects, among which, some expired sensors will cause missing values of the corresponding features \cite{hou2017one}.

Given that the feature values are severely missing, the performance of a classification model trained on such a dataset will be significantly degenerated. It is thus important to recover the missing values. The most reliable way is to acquire the ground-truth values for the missing features. Unfortunately, acquiring a feature value usually involves special devices or complex processes, leading to high acquisition costs. Nevertheless, features are often correlated with each other, and redundant information is contained across features. Thus it may not be necessary to query all feature values. Instead, one can query a part of the features, and then recover the others from the observed entries.

Matrix completion would be a useful tool for recovering missing entries of the feature matrix, which has been extensively studied \cite{CY14,FK15,DBLP:conf/sdm/ZengLCXZL15,DBLP:journals/kais/SunGLX17}. However, existing approaches neglect the class labels, which may provide supervised information to guide the matrix completion to a desired solution. In practice, the observed entries may be noisy, and are not adequate to provide sufficient information to recover the missing values. Especially when the missing rate is high, there could be a large number of possible matrices that can well fit the observed values. The class labels, which strongly depend on the feature representations, are expected to narrow the choice over all possible matrices.

Furthermore, different features may have different contributions to recovering the missing values as well as improving the classification model. Some features are crucial while others may be less important. It is thus practical to actively select the most informative features to acquire their ground-truth values, and recover the missing values based on the observed features.

Traditional active learning algorithms select the most informative unlabeled instances to query their labels, and can significantly reduce the annotation cost \cite{SB12,HJZ14}. Similar ideas have been extended in order to perform active feature acquisition \cite{RN15,BAG17,CD17}. These methods typically try to estimate the expected utility of a feature value for improving the model performance, and then query the ground-truth value for the feature with maximum expected utility. However, some features with high potential utility can be recovered by matrix completion, and thus querying their values can be waste of acquisition costs.

In this paper, we jointly perform active feature querying and supervised matrix completion to minimize the acquisition cost. To exploit the label information for effective matrix completion, we propose an objective function that consists of the reconstruction error, the low rank regularizer and the empirical classification error. By minimizing this objective function, the recovered feature matrix is expected to on one hand well fit the structure in feature space, and on the other hand follow the label supervision to be discriminative. To select the most informative entry for active feature acquisition, we propose a variation based criterion, which estimates the informativeness of a feature value on recovering the missing values as well as improving the classification model. Furthermore, we introduce a bi-objective optimization method to handle the case where the acquisition cost varies for different features.

Theoretical analysis is presented to give an upper bound on the reconstruction error of the proposed matrix completion algorithm. Further, experiments are performed on different datasets to validate the effectiveness of the proposed approach. Results demonstrate that our approach can recover the matrix accurately, and achieve effective classification with less feature acquisition cost.

The rest of the paper is organized as follows: we review related works in Section 2, and introduce the proposed approach in Section 3. Section 4 presents the settings and results of the experiments, followed by the conclusion in Section 5.

\section{Related Work}

Active learning has been widely studied for reducing the labeling cost~\cite{SB12,HJZ14}. Classical studies focused on designing a selection criterion such that selected instances can improve the model maximally. Informativeness is one of the most commonly used criteria, which estimates the ability of an instance in reducing the uncertainty of a statistical model. Typical techniques for informative sampling include statistical methods~\cite{CD95}, SVM-based methods~\cite{ST01}  and query-by-committee methods~\cite{YF97}, etc.

Differently from traditional active learning that targets reducing the labeling cost, there is another branch of research employing similar ideas to reduce the feature acquisition cost~\cite{YTDC15}. These methods iteratively query the ground-truth values for the actively selected features, and are expected to improve the learning performance with least queries. Some methods tried to estimate the expected utility of each feature to improve the model, and then select the top features with maximum expected utility to query their values. For example, in~\cite{PF05}, a criterion was proposed to estimate the expected improvement of accuracy per unit cost, and then the most cost-effective feature values were iteratively acquired. A similar approach was proposed in \cite{VB08}, where the learning task is clustering instead of classification, and thus the corresponding criterion estimates the expected improvement in clustering quality per unit cost. There is another category of methods called instance completion. Instead of querying one specific feature value, they selected a small batch of incomplete instances with missing features, and queried all missing values for the selected instances each time. The instances are actively selected aiming to improve the classification performance. For example, the authors of~\cite {SK13} proposed to estimate the expected utility of each instance for active selection, and also derived a probabilistic lower bound on the error reduction achieved with the proposed technique. The method in \cite{DA15} chose the top $k$ instances based on a derived upper bound on the expected distance between the next classifier and the final classifier.

A common limitation of these methods is that they do not consider the case where some of the missing features can be accurately recovered from the observed entries, and thus may waste the acquisition cost of unnecessary queries.  There is one study that tried to query both missing features and labels, and built an imputation model for missing features~\cite{MS14}. However, it requires a complete set of training examples for training a model, which may not be satisfied in real applications .

Matrix completion is a classical approach for recovering the missing entries of a partially observed matrix. It has been successfully applied to collaborative filtering \cite{DBLP:conf/icml/RennieS05}, dimensionality reduction \cite{DBLP:journals/ijcv/WeinbergerS06}, multi-class/multi-label learning \cite{AG10, CR15}, clustering \cite{DBLP:journals/corr/abs-1112-5629,DBLP:conf/icdm/YiYJJM12}, etc. One main category of existing methods is statistical matrix completion based on the low-rank assumption \cite{CE09,CY14,KR10,WZ12,JP13,NS12}. These methods usually transform the matrix completion task into an optimization problem, and try to find a low-rank matrix to fit the observed entries. There are some structural matrix completion methods which explicitly analyze the information contained in the observed entries and are capable of evaluating whether the observations are theoretically  sufficient for recovering the missing values \cite{MR09,AS10,FK15}.

In some cases the observed entries are not enough to recover the others, and thus further queries are needed to acquire more ground-truth values for some missing entries. Given this background, there are some active learning approaches proposed to query the most informative entries for completion \cite{RN15}. For example, a general framework was proposed in \cite{CS13} for active matrix completion, where existing matrix completion methods can be enhanced with an uncertainty sampling strategy. In \cite{SD13}, the authors firstly estimated the posterior distribution with variational approximations or Markov chain Monte Carlo sampling, and then queried the entries for collaborative prediction. The algorithm in \cite{RN15} unified active querying and matrix completion in a single framework. There are some other approaches which study active completion with specific requirements on the matrix \cite{BAG17,CD17}.

While all the above studies are not theoretically grounded, there are two works focusing on adaptive querying for matrix completion with theoretical results. One is~\cite{DBLP:conf/nips/KrishnamurthyS13} which firstly sampled several rows, and adaptively decided which columns are need to be fully observed. The other is~\cite{BAG17} which actively completed a low-rank positive semi-definite matrix. Although these two works are theoretically sound, they do not consider any supervision information.
%active feature extraction
%bi-objective optimization

\section{The Proposed Approach}

We denote by $D=\{(\x_i, y_i)\}$ a  dataset with $n$ instances, where $\x$ is a $d$-dimensional real feature vector for the $i$-th instance and $y_i$ is its class label. Let $\X\in\mathbb{R}^{n\times d}$ be the ground-truth feature matrix of the $n$ instances, where each column represents one dimensionality of the $d$-dimensional feature space. Here we consider the feature missing problem, where $\X$ is only partially observed. We denote by $\Omega$ the set of indices for the observed entries of $\X$. In the rest of this section, we will firstly propose a supervised matrix completion method, and then present an active feature acquisition approach.

\subsection{Supervised matrix completion}\label{sec:3.1}
We focus on the matrix completion problem under the supervised classification setting, where the task is to learn a function $f$ for predicting the class labels of instances. Matrix completion is a challenging problem because observed entries are usually limited, and often do not contain sufficient information for recovering missing values. Since there are an arbitrary number of possible matrices that perfectly match the observed entries, external knowledge is needed to find the optimal one closest to the ground-truth. Low-rank is a common assumption for matrix completion, which exploits the structure information in the feature space. In this paper, we further exploit the supervised information contained in class labels to guide the matrix completion to a desirable solution. Classification function $f$ is a mapping from the feature space to the label space, and thus can be utilized to inversely transfer the label information for feature recovering. For example, given an instance with missing features and its class label, we denote by $\x$ a recovered feature vector. Assuming the classifier $f$ is reliable, if the  prediction $f(\x)$ is faraway from the ground-truth label $y$, then it is more likely that feature vector $\x$ is not accurately recovered. Based on this motivation, we propose to minimize the empirical classification error along with the reconstruction error and the matrix rank within one unified framework, where the feature matrix and the classification model are alternately optimized.

On one hand, we want to accurately recover the ground-truth feature matrix from the partial observation of $\X$ with the low-rank assumption. On the other hand, the classification model $f$, which is trained with the recovered matrix $\widehat{\mathbf{X}}$, is expected to have a small empirical error. Based on this argument, we define our objective function as follows.
\begin{equation}\label{eq:obj}
\min_{\bhX,f} \frac{1}{2}\|\mathcal{R}_{\Omega}(\bhX-\X)\|^2_\mathrm{F} + \lambda_1\|\bhX\|_\mathrm{tr} + \lambda_2\sum_{i=1}^n \ell(y_i, f(\hat{\x}_i)),
\end{equation}
where $\mathcal{R}_\Omega : \mathbb{R}^{n\times d}\rightarrow \mathbb{R}^{n\times d}$, 
\begin{equation*}
\left[\mathcal{R}_\Omega(X)\right]_{i,j}=\Big\{
\begin{aligned}
&X_{i,j}&\text{if } (i,j)\in\Omega ,\\
&0&\text{otherwise},
\end{aligned}
\end{equation*}
$\|\cdot\|_\mathrm{tr}$ is the trace norm, $\|\cdot\|_\mathrm{F}$ is the Frobenius norm, and $\lambda_1,\lambda_2\ge0$ are regularization parameters.

We assume that the loss function $\ell$ can be written as a function parameterized by $\bhX$, and it is Lipschitz smooth with respect to $\bhX$. One example is the linear classifier with the squared loss, i.e., $f(\x_i)=\bm{w}^\top\x_i$, where $\bm{w}\in \mathbb{R}^d$; then we have $\sum_{i=1}^n \ell(y_i, f(\x_i))=\|\X\bm{w}-\bm{y}\|^2$, where $\bm{y}=[y_1, y_2, \ldots, y_n]$ and $\|\cdot\|$ denotes the $\ell_2$ norm. In the following, we will write $\sum_{i=1}^n\ell(y_i, f(\x_i))$ as $\ell(\X, f)$ for notational simplicity. Then the optimization problem becomes
\begin{equation}
\min_{\bhX,f} \frac{1}{2}\|\mathcal{R}_{\Omega}(\bhX-\X)\|^2_{\mathrm{F}} + \lambda_1\|\bhX\|_{\mathrm{tr}} + \lambda_2\ell(\bhX, f),
\end{equation}
which can be solved by alternately optimizing $\bhX$ and $f$.

When optimizing $\bhX$ with fixed $f$, we have
\begin{align}
\min_{\bhX} \frac{1}{2}\|\mathcal{R}_{\Omega}(\bhX-\X)\|^2_{\mathrm{F}} + \lambda_1\|\bhX\|_{\mathrm{tr}} + \lambda_2\ell(\bhX).
\end{align}
We will exploit the accelerated proximal gradient descend~\cite{tseng2008accelerated} which is a classical optimization technique in trace norm minimization to solve this problem. Let
\begin{equation*}
g(\bhX)=\frac{1}{2}\|\mathcal{R}_{\Omega}(\bhX-\X)\|^2_{\mathrm{F}} + \lambda_2\ell(\bhX),
\end{equation*}
and
\begin{equation*}
h(\bhX, \Z)=g(\Z)+\left<\triangledown g(\Z), \bhX - \Z\right>+\lambda_1\|\bhX\|_{\mathrm{tr}},
\end{equation*}
where
\begin{equation*}
\triangledown g(\Z)=\mathcal{R}_\Omega(\Z-\bhX)+\lambda_2\frac{\partial\ell}{\partial \Z}.
\end{equation*}
We summarize the main steps here:
\begin{itemize}
	\item Choose $\theta_0=\theta_{-1}\in (0,1]$, $L>1$, $\bhX_0=\bhX_{-1}$, $\gamma>1$. Set $k=0$.
	\item In the $k$-th iteration,	
	\begin{itemize}
		\item Set $\Z_k=\bhX_k+\theta_k(\theta_{k-1}^{-1}-1)(\bhX_k-\bhX_{k-1})$.
		\item Set $\bhX_{k+1}=\argmin_{\bhX}\left\{h(\bhX,\Z_k)+\frac{L}{2}\|\bhX-\Z_k\|_{\mathrm{F}}^2\right\}$.
		\item While $g(\bhX_{k+1})+\lambda_1\|\bhX_{k+1}\|_{\mathrm{tr}}>h(\bhX_{k+1},\Z_k)+\frac{L}{2}\|\bhX_{k+1}-\Z_k\|_{\mathrm{F}}^2$:
		\begin{itemize}
			\item Increase $L=\gamma L$.
			\item Update $\bhX_{k+1}=\argmin_{\bhX}\left\{h(\bhX,\Z_k)+\frac{L}{2}\|\bhX-\Z_k\|_{\mathrm{F}}^2\right\}$.
		\end{itemize}
		\item Set $\theta_{k+1}=\sqrt{\theta_k^4+4\theta_k^2}-\theta_k^2/2$.
		\item Update $k=k+1$.
	\end{itemize}	
\end{itemize}
The iteration continues until convergence. In the above steps, we have not specified how to obtain $\bhX_{k+1}$ and next we will explain this. We rewrite the problem as
\begin{equation}
\min_{\bhX} \left<\triangledown g(\Z_k), \bhX-\Z_k\right>+\frac{L}{2}\|\bhX-\Z_k\|_{\mathrm{F}}^2+\lambda_1\|\bhX\|_{\mathrm{tr}},
\end{equation}
which is equivalent to
\begin{equation}
\min_{\bhX} \frac{L}{2}\left\|\bhX-\left(\Z_k-\frac{1}{L}\triangledown g(\Z_k)\right)\right\|_{\mathrm{F}}^2+\lambda_1\|\bhX\|_{\mathrm{tr}}.
\end{equation}
This can be solved by Singular Value Thresholding (SVT)~\cite{CCS10}, which performs singular value decomposition on $\Z_k-\frac{1}{L}\triangledown g(\Z_k)=\U\bSigma \V^\top$. Let $\widehat{\Sigma}_{ii}=\max(0,\Sigma_{ii}-\frac{\lambda_1}{L})$, the solution is given by $\U\widehat{\bSigma}\V^\top$.

Finally, the classification model $f$ is optimized with fixed $\bhX$, which can be efficiently solved using existing algorithms. These two procedures are repeated until convergence.
%The pseudo code of the algorithm for supervised matrix completion is summarized in Algorithm \ref{alg:SMC}.

\subsection{Active feature acquisition}
In this subsection, we discuss how to actively query the ground-truth values as most informative features, with the target of improving the model mostly based on the smallest number of queries. We will first present a novel criterion for estimating the informativeness of a feature, and then introduce a method to handle the case where the acquisition cost varies for different features.

\subsubsection{Variance-based selection}
In traditional active learning, if the model is less certain about the prediction on an instance, then the instance is considered to be more informative for improving the model, and will be more likely to be selected for label querying~\cite{HJZ14}. Inspired by this idea, we also propose an uncertainty criterion to estimate the informativeness of a feature. The challenge here is that the informativeness should reflect the usefulness of a feature both for recovering other entries and for training the classification model. Notice that the objective function defined in Eq. (\ref{eq:obj}) does consider the two aspects simultaneously. At each iteration of active learning, after a small batch of feature values is acquired, the algorithm in Section~\ref{sec:3.1} will be employed to optimize Eq. (\ref{eq:obj}) for matrix completion. The output of the matrix completion may vary from iteration to iteration. If the variance of an entry over iterations is large, it implies that the entry can not be certainly decided by the algorithm, and thus may contain more useful information to recover the feature matrix and optimize the classification model. Denoting by $\X^t$ the completed matrix at the $t$-th iteration, the informativeness of the $j$-th feature of $\x_i$ is defined as:
\begin{equation}\label{eq:info}
I_{i,j}=\sum_{t=1}^T(X_{i,j}^t-\bar{X}_{i,j})^2,
\end{equation}
where $\bar{X}_{i,j}=\frac{1}{T}\sum_{t=1}^T X_{i,j}$ is the mean value of $X_{i,j}$ over all iterations. Then a small batch of most uncertain features with largest informativeness is selected to query their ground-truth values. The pseudo code of the algorithm for active feature acquisition is summarized in Algorithm \ref{al:AFA}. We call the proposed algorithm Active Feature Acquisition Supervised Matrix Completion (AFASMC).

\begin{algorithm}[!tb]
	\caption{The AFASMC algorithm}\label{al:AFA}
	\label{algorithm}
	\begin{algorithmic}[1]
		\STATE \textbf{Input:}
		\STATE \quad $D$: the data set of $n$ instances, with $\bhX$ as the feature matrix
		\STATE \quad $\Omega$: the set of indices for observed entries
		\STATE \quad $\lambda_1$, $\lambda_2\ge 0$: the parameters
		\STATE \textbf{Process:}
		\begin{ALC@g}
			\STATE \textbf{For:} $t=1:T$
			\begin{ALC@g}
				\STATE \textbf{Repeat}
				\begin{ALC@g}
					\STATE recover the matrix $\bhX$ with fixed classification model $f$
					\STATE optimize the model $f$ with fixed $\bhX$
				\end{ALC@g}
				\STATE \textbf{Until} convergence
				\STATE \textbf{For} each missing entry $\hX_{i,j}$		
				\begin{ALC@g}
					\STATE	calculate the average value over the $t-1$ iterations
					\STATE	calculate the informativeness of $\hX_{i,j}$ as Eq. (\ref{eq:info})
				\end{ALC@g}
				\STATE \textbf{End For}
				\STATE select a batch of entries with maximum informativeness
				\STATE query the ground-truth values for the selected entries
				\STATE set the corresponding elements of $\Omega$ to 1
			\end{ALC@g}
			\STATE \textbf{End For}
		\end{ALC@g}
	\end{algorithmic}
\end{algorithm}

Note that it is not necessary to calculate the variance based on all iterations. Generally speaking, it is more important to capture the change of an entry within recent iterations. For example, if an entry has a large variance at an early stage, but becomes stable after a few queries, it implies that this entry may have been well recovered from the recently acquired features, and thus does not need to be queried any more. We will discuss this in the experiments in more detail.

%[optional: maybe a figure to show the variance.]

\subsubsection{Cost-aware selection}
Finally, we discuss a more complicated case, where the cost of acquiring a feature value varies for different features. This is a common case in real applications. For example, it is much more costly to perform an fMRI scan than blood examination for diagnosing a patient. While there is typically a conflict between the informativeness and acquisition cost of a feature, we propose to balance these two factors for achieving the best cost-effectiveness. We denote the cost for acquiring the $j$-th dimension of the features by $C_j$. Note here we assume that the acquisition cost is independent of the instance. We offer two optional strategies to consider the acquisition cost. The most straightforward method is to simply divide the informativeness by the acquisition cost. So we can have the selection strategy as:
\begin{equation}
\argmax_{(i,j)\notin \Omega}\frac{I_{i,j}}{C_j}.
\end{equation}
This strategy provides a simple solution for cost-aware selection, but may fail when one of the two factors dominates the other.

In what follows, we introduce another solution by bi-objective optimization. In each iteration of our algorithm, we select a small batch of missing entries of the feature matrix to acquire their ground-truth values. This is a typical subset selection problem. Generally, a subset selection problem tries to select a subset $S$ from a large set $V$ with an objective function $\mathcal{J}$ and a constraint of the subset size. It can be formalized as
\begin{equation}
\label{eqn:subset}
\argmin_{S\subseteq V} \mathcal{J}(S) \qquad \mathrm{s.t.}\quad |S|\leq b,
\end{equation}	
where $ |\cdot| $ denotes the size of a set, and $b$ is the maximum number of selected elements. Further, for convenience of presentation, the subset selection problem is reformulated as optimizing a binary vector. We introduce a binary vector $ s\in\{0,1\}^{n} $ to indicate the subset membership, where $ s_{i}=1 $ if the $ i$-th element in $ V$ is selected, and $ s_{i}=0 $ otherwise. Following the method in \cite{qian2015subset}, the subset selection problem in Eq. (\ref{eqn:subset}) can be written as a bi-objective minimization problem:
\begin{align}
&\qquad\argmin_{s\in\left\lbrace 0,1\right\rbrace ^{n}} (\mathcal{J}_{1}(s), \mathcal{J}_{2}(s)),\\
&\mathcal{J}_{1}(s)=\Big\{
\begin{array}{rll}
+\infty  &\mbox{if } s=\{0\}^{n} \text{ or } \,|s|\geq2b, \\
\mathcal{J}(s) &\text{otherwise}.
\end{array} \quad \mathcal{J}_{2}(s)=|s|,\nonumber
\end{align}
where $|s|$ denotes the number of 1s in $s$. Obviously, the problem is for sparse selection with the target of minimizing $\mathcal{J}$. Here $\mathcal{J}_1$ is set to $+\infty$ to avoid trivial solutions or over-sized subsets. In our case, we want to maximize the informativeness in Eq. (\ref{eq:info}), and at the same time minimize the acquisition cost of the selected entries. We thus can redefine the two objective functions $\mathcal{J}_1$ and $\mathcal{J}_2$ correspondingly, and have the following bi-objective optimization problem.
\begin{align}\label{eq:final}
&\argmin_{s\in\left\lbrace 0,1\right\rbrace ^{n}} (\mathcal{J}_{1}(s), \mathcal{J}_{2}(s)),\\
&\mathcal{J}_{1}(s)=\Big\{
\begin{array}{ll}
+\infty&  \mbox{if } s=\{0\}^{n} \text{ or } \mathcal{J}_{2}(s)\geq2b, \\
-\sum_{ij} s(i,j)\cdot I_{ij}& \text{otherwise}.
\end{array}\nonumber\\
&\mathcal{J}_{2}(s)=\sum_{ij} s(i,j)\cdot C_{j}.\nonumber
\end{align}
Here $b$ is the budget for the acquisition cost in each iteration, and $s(i,j)$ is used to denote the element of $s$ corresponding to the entry of the $i$-th row and $j$-th column in matrix $X$. Again, $\mathcal{J}_1(s)$ is set to $+\infty$ to exclude trivial or over-cost solutions. We employ a recently proposed Pareto optimization algorithm called Pareto Optimization for Subset Selection (POSS) \cite{qian2015subset} to solve this problem. POSS is an evolutionary style algorithm, which maintains a solution archive, and iteratively update the archive by replacing some solutions with better ones. In detail, it initializes the archive with a solution of empty subset selection. In each iteration, a solution $s$ is selected from the current archive, and a new solution $s'$ is generated by randomly flipping bits of $s$. The two objective values $\mathcal{J}_1(s')$ and $\mathcal{J}_2(s')$ are then computed to compare $s'$ with the archived solutions. Specifically, if there exists one solution $s$ in the achieve that satisfies both the following conditions:
\begin{align*}
&\mathcal{J}_1(s)\leq \mathcal{J}_1(s')\text{ and } \mathcal{J}_2(s)\leq \mathcal{J}_2(s'),\\
&\mathcal{J}_1(s)< \mathcal{J}_1(s') \text{ or } \mathcal{J}_2(s)< \mathcal{J}_2(s'),
\end{align*}
then $s'$ will be ignored; otherwise, $s'$ will be added to the solution archive, and at the same time all the archived solutions $s$ that satisfy
\begin{equation*}
\mathcal{J}_1(s')\leq \mathcal{J}_1(s)\text{ and } \mathcal{J}_2(s')\leq \mathcal{J}_2(s)
\end{equation*} will be removed from the solution archive. This process is repeated until reaching a specified number of iterations. At last, the best solution with the minimal value on $J_1$ and within the cost budget will be selected as the final solution.

\subsection{Theoretical analysis}
In this subsection, we will present a theoretical bound on the reconstruction error of the supervised matrix completion method introduced in Section 3.1. For the loss between $\X\w$ and $\y$, i.e., the term $\sum_{i=1}^n\ell(y_i,f(\x_i))$ in Eq. (\ref{eq:obj}), here we discuss a more strict case by enforcing $\X\w$ and $\y$ to be equal. It is reasonable to relax this strict constraint as in Eq. (\ref{eq:obj}) to cope with possible noises. The relaxation is also benefited by more flexible choice of the loss function, for example $\|\X\w-\y\|$, as well as ease of optimization. For convenience of presentation, we rewrite the noiseless counterpart of Eq. (\ref{eq:obj}) as:
\begin{eqnarray}\label{eqn:opt-obj}
\min_{\bhX} \sum_{(i,j)\in\Omega} (\hX_{i,j}-X_{i,j})^2\ \ \text{ s.t. }\ \|\bhX\|_{\mathrm{tr}}^2\le \beta\sqrt{rnd}, \ f(\bhX)=\y,
\end{eqnarray}
where $\beta$ and $r$ are constants. We assume $\bhX^*$ is the optimal solution for Eq. (\ref{eqn:opt-obj}), and try to analyze the difference between the solution of our algorithm and the optimal solution. Before discussing the property of the solution, we first define the \emph{coherence} of a matrix, which will be used later.
\begin{definition}
	For a rank-$r$ matrix $\M\in\R^{n\times m}$ whose SVD is $\M=\U\bSigma\V^\top$, we use the following value as the \emph{coherence},
	\begin{eqnarray*}
		\mu(\M)=\max\{\max_{1\le i\le n}\| \U_{i,*}\|, \max_{1\le j\le m}\|\V_{j,*}\|\},
	\end{eqnarray*}
	where $\U_{i,*}$ $(\V_{j,*})$ denotes the $i$th $(j$th$)$ row of $\U$ $(\V)$.
\end{definition}

Note that compared to~\cite{CE09,DBLP:conf/nips/XuJZ13}, for ease of use, in this paper we do not normalize the coherence by the size of the matrix. Coherence measures how the values of the entries are distributed in a matrix. The lower the coherence is, the more average the values of the entries are distributed. Apparently, if there is no entry that has a ``peak'' value in a matrix, the matrix is easier to be completed with partial observations. Based on this definition, we give our theoretical results in Theorem \ref{thm:thm1}.

\begin{thm}\label{thm:thm1}
	Suppose that  $\|\X\|^2_{\mathrm{tr}}\le \beta\sqrt{rnd}$, $f(\X)=\y$ and $\Omega$  is chosen independently at random following a binomial model with probability $|\Omega|/(nd)$. Let $\bhX^*$ be the solution to the optimization problem Eq. (\ref{eqn:opt-obj}) and $\mu=\max_{\bhX\in G}\mu(\bhX)$, where $G\subset\R^{n\times d}$ is $G=\left\{\bhX\in\R^{n\times d}\left\vert\right. \|\bhX\|^2_{\mathrm{tr}}\le \beta\sqrt{rnd},f(\bhX)=\y\right\}$	for some $r\le \min\{n,d\}$, and $\beta\ge 0$. Then with probability at least $1-C/(n+d)$, we have
	\begin{eqnarray*}
		\frac{1}{nd}\|\bhX^*-\X\|^2_{\mathrm{F}}\le 2
		\left(C_0\mu^2\beta \sqrt{\frac{r(n+d)}{|\Omega|}}\sqrt{1+\frac{(n+d)\log(n+d)}{|\Omega|}}\right).
	\end{eqnarray*}
\end{thm}
Theorem \ref{thm:thm1} provides an upper bound of the reconstruction error for the proposed supervised matrix completion algorithm. Moreover, it is obvious that a smaller upper bound can be expected by increasing $|\Omega|$.  This also motivates us to iteratively acquire more feature values. Below we present a sketch of proof of Theorem \ref{thm:thm1}. A detailed proof is available in a longer version on arXiv~\cite{DBLP:journals/corr/abs-1802-05380}.%\footnote{https://arxiv.org/abs/1802.05380}.

To prove Theorem \ref{thm:thm1}, we firstly define $\L_{\Omega}(\bhX)=-\sum_{(i,j)\in\Omega} (\hX_{i,j}-X_{i,j})^2$ and $\bL_{\Omega}(\bhX)=\L_{\Omega}(\bhX)-\L_{\Omega}(\X)$. Note that because $\X$ is a constant matrix, subtracting $\L_{\Omega}(\X)$ will not affect optimization of the objective function, i.e., they will both have the same $\bhX^*$ leading to the optimum. Then we will use the following three lemmas to prove Theorem~\ref{thm:thm1}:
\begin{lemma}\label{lem:main}
	Assume that $\|\X\|^2_{\mathrm{tr}}\le \beta\sqrt{rnd}$, and $\mu=\max_{\bhX\in G}\mu(\bhX)$, then with probability at least $1-C/(n+d)$ we have
	\begin{eqnarray*}
\sup_{\bhX\in G}\left\|\bL_{\Omega}(\bhX)-\E[\bL_{\Omega}(\bhX)]\right\|\le \left(C_0\mu^2\beta \sqrt{r}\sqrt{|\Omega|(n+d)+nd\log(n+d)}\right),
	\end{eqnarray*}
	where the expectation is over the choice of $\Omega$.
\end{lemma}

We can also easily derive the following result from~\cite{DBLP:journals/corr/abs-1209-3672}:
\begin{lemma}\label{lem:lem2}
	If $E\in \R^{d_1\times d_2}$, and each entry $E_{i,j}$ is a Radamacher random variable; $\Delta\in \{0,1\}^{d_1\times d_2}$, and each entry $\Delta_{i,j}$ is independently sampled when $\Delta_{i,j}=1$ with probability $n/(d_1d_2)$ and $0$ with probability $1-n/(d_1d_2)$. Then we have
	\begin{eqnarray*}
		\E\left[\left\|E\circ \Delta\right\|^h\right]\le C2^h (1+\sqrt{6})^h \left(\frac{n(d_1+d_2)+d_1d_2\log(d_1+d_2)}{d_1d_2}\right)^{h/2},
	\end{eqnarray*}
	provided that $h\ge 1$ and $C$ is a constant.
\end{lemma}

Further, the trace norm of the Hadamard product of two matrices is bounded as follows:
\begin{lemma}\label{lem:lem3}
	Assume that there are two matrices $\A$ and $\B$ that have the same shape, then we have $\|\A\circ \B\|_{\mathrm{tr}}\le \mu(\A)^2 \|\A\|_{\mathrm{tr}}\|\B\|_{\mathrm{tr}}$, where $\circ$ is the Hadamard product.
\end{lemma}

\begin{table*}[!t]
	\centering
	\caption{The comparison results on matrix completion. The reconstruction error as well as classification accuracy are reported with 60\% and 80\% entries observed respectively.} \label{table:mc}
	\renewcommand{\multirowsetup}{\centering}
	\begin{tabular}{c | c  | c c c c | c c c c}
		\hline
		Data  & Observed Rate & AFASMC & OptSpace &LmaFit  & NNLS &  AFASMC & OptSpace &LmaFit  & NNLS  \\
		\hline
		\multicolumn{2}{l}{}& \multicolumn{4}{l}{Reconstruction Error} & \multicolumn{4}{l}{Test Accuracy (\%)}\\
		\hline
		\multirow{3}{*}{abalone}
		& 60\%  &\textbf{ 0.13$\pm$0.01 }& 0.38$\pm$0.01 & 0.14$\pm$0.00 & 0.14$\pm$0.00 & \textbf{78.5$\pm$1.2 }& 71.8$\pm$0.8 & 78.3$\pm$1.3 & 78.4$\pm$1.2 \\
		& 80\%  & \textbf{0.07$\pm$0.00 }& 0.23$\pm$0.01 & 0.09$\pm$0.00 &\textbf{ 0.07$\pm$0.00}& \textbf{79.7$\pm$0.7} & 76.3$\pm$1.2 & 79.5$\pm$0.6 & 79.5$\pm$0.9 \\
		\hline
		\multirow{3}{*}{letter}
		& 60\%  & \textbf{0.18$\pm$0.00 }& 0.33$\pm$0.01 & 0.24$\pm$0.00 & 0.23$\pm$0.00 & \textbf{98.5$\pm$0.6} & 92.1$\pm$3.7 & 97.0$\pm$1.3 & 94.3$\pm$1.0 \\
		& 80\%  & \textbf{0.11$\pm$0.00 }& 0.29$\pm$0.00 & 0.17$\pm$0.00 & 0.17$\pm$0.01 & \textbf{99.3$\pm$0.2 }& 94.2$\pm$0.8 & 99.1$\pm$0.4 & 94.6$\pm$1.1 \\
		\hline
		\multirow{3}{*}{image}
		& 60\%  &\textbf{ 0.25$\pm$0.00} & 0.54$\pm$0.01 & 0.36$\pm$0.05 & 0.56$\pm$0.03 & \textbf{79.3$\pm$2.0} & 67.1$\pm$2.2 & 75.5$\pm$2.1 & 70.1$\pm$3.5 \\
		& 80\%  &\textbf{ 0.16$\pm$0.00 }& 0.51$\pm$0.00 & 0.18$\pm$0.00 & 0.23$\pm$0.08 & \textbf{81.0$\pm$2.3 }& 68.1$\pm$1.5 & 79.8$\pm$2.5 & 80.5$\pm$3.2 \\
		\hline
		\multirow{3}{*}{chess}
		& 60\%  & \textbf{0.43$\pm$0.00} & 1.57$\pm$0.01 & 0.48$\pm$0.00 & 1.28$\pm$0.02 &\textbf{ 94.3$\pm$0.9 }& 52.1$\pm$1.8 & 93.3$\pm$0.8 & 53.8$\pm$1.7 \\
		& 80\%  & \textbf{0.29$\pm$0.00 }& 1.63$\pm$0.01 & 0.33$\pm$0.00 & 1.21$\pm$0.01 & \textbf{94.8$\pm$0.4} & 52.8$\pm$1.9 & 94.7$\pm$0.7 & 77.9$\pm$3.6 \\
		\hline
		\multirow{3}{*}{HillValley}
		& 60\%  & 0.04$\pm$0.00 & 0.06$\pm$0.00 & 0.04$\pm$0.00 & \textbf{0.03$\pm$0.00}&\textbf{ 51.5$\pm$3.2} & 48.7$\pm$3.5 & 49.4$\pm$3.7 & 50.7$\pm$2.8 \\
		& 80\%  & \textbf{0.03$\pm$0.00} & 0.06$\pm$0.00 & \textbf{0.03$\pm$0.00} &\textbf{ 0.03$\pm$0.00} &\textbf{ 51.1$\pm$2.1} & 49.5$\pm$3.3 & 50.8$\pm$2.3 & 49.7$\pm$2.0 \\
		\hline
		\multirow{3}{*}{HTRU2}
		& 60\%  & \textbf{0.29$\pm$0.00} & 0.59$\pm$0.00 & 0.63$\pm$0.00 & \textbf{0.29$\pm$0.00} & \textbf{97.2$\pm$0.2}& 90.8$\pm$0.3 & 97.1$\pm$0.2 & 97.1$\pm$0.2 \\
		& 80\%  & \textbf{0.16$\pm$0.00 }& 0.39$\pm$0.00 & 0.45$\pm$0.00 & 0.17$\pm$0.00 &\textbf{ 97.4$\pm$0.2}& 94.3$\pm$0.2 & 97.3$\pm$0.2 &\textbf{ 97.4$\pm$0.2} \\
		\hline
	\end{tabular}
\end{table*}

The proof of the lemmas is available in a longer version at arXiv~\cite{DBLP:journals/corr/abs-1802-05380}. Next we show how from Lemma~\ref{lem:main} we can derive our Theorem~\ref{thm:thm1}. Note that for any choice of matrix $\A\in G$, we have
\begin{eqnarray*}
	&&\E\left[\bL_{\Omega}(\A)-\bL_{\Omega}(\X)\right]=\E\left[\L_{\Omega}(\A)-\L_{\Omega}(\X)\right]\\
	&=&\E\left[-\sum_{(i,j)\in\Omega} (A_{i,j}-X_{i,j})^2\right]=-\frac{|\Omega|}{nd}\left(\sum_{(i,j)} (A_{i,j}-X_{i,j})^2\right),
\end{eqnarray*}
where the expectation is over $\Omega$. We can also have
\begin{eqnarray*}
	&&\bL_{\Omega}(\A)-\bL_{\Omega}(\X)=\bL_{\Omega}(\A)-\E\left[\bL_{\Omega}(\A)\right]-\bL_{\Omega}(\X)\\
	&&+\E\left[\bL_{\Omega}(\X)\right]+\E\left[\bL_{\Omega}(\A)-\bL_{\Omega}(\X)\right]\\
	&\le&2\sup_{\A\in G}\left\|\bL_{\Omega}(\A)-\E[\bL_{\Omega}(\A)]\right\|
	-\frac{|\Omega|}{nd}\left(\sum_{(i,j)} (A_{i,j}-X_{i,j})^2\right).
\end{eqnarray*}

Replacing $\A$ by $\bhX^*$ which is the optimal solution to Eq. (\ref{eqn:opt-obj}) and noting that $\bL_{\Omega}(\bhX^*)\ge\bL_{\Omega}(\X)$, we have
%\begin{eqnarray*}
%	2\sup_{\A\in G}\left\|\bL_{\Omega}(\A^*)-\E[\bL_{\Omega}(\A)]\right\|
%	-\frac{|\Omega|}{nd}\left(\sum_{(i,j)} (A_{i,j}-X_{i,j})^2\right)\ge 0,
%\end{eqnarray*}
%i.e.,
\begin{eqnarray*}
	\frac{|\Omega|}{nd}\left(\sum_{(i,j)} (A_{i,j}-X_{i,j})^2\right)\le 2\sup_{\A\in G}\left\|\bL_{\Omega}(\A)-\E[\bL_{\Omega}(\A)]\right\|.
\end{eqnarray*}

Using Lemma~\ref{lem:main}, with probability at least $1-\frac{C}{n+d}$, we have
\begin{eqnarray*}
	\frac{|\Omega|}{nd}\|\bhX^*-\X\|^2_{\mathrm{F}}\le 2
	\left(C_0\mu^2\beta \sqrt{r}\sqrt{|\Omega|(n+d)+nd\log(n+d)}\right).
\end{eqnarray*}
Further applying $\sqrt{nd}\le n+d$, we have
\begin{eqnarray*}
	\frac{1}{nd}\|\bhX^*-\X\|^2_{\mathrm{F}}\le 2
	\left(C_0\mu^2\beta \sqrt{\frac{r(n+d)}{|\Omega|}}\sqrt{1+\frac{(n+d)\log(n+d)}{|\Omega|}}\right).
\end{eqnarray*}

\section{Experiments}
In this section, we experimentally investigate the proposed method.
\subsection{Settings}
We perform experiments on 6 benchmarks, namely  \emph{abalone}, \emph{letter}, \emph{image}, \emph{chess}, \emph{HillValley} and \emph{HTRU2}. The number of entries in the matrix varies from 22,960 to 143,184. For each dataset, we randomly separate the set into two subsets, one with 70\% examples for training, and the other one with 30\% examples for testing. We repeat the random partition 10 times and report the average results.

%\begin{table}[!tb]
%	\centering
%	\caption{The information of the datasets used in the experiments.}\label{table:data}
%	\renewcommand{\multirowsetup}{\centering}
%	\begin{tabular}{c|c|c|c}
%		\hline
%		Dataset & \# Instances &\# Features & Matrix Size \\
%		\hline
%		abalone&2870 &8 & 22960\\
%		\hline
%		letter      &1555 & 16&24880\\
%		\hline
%		image	&2086 &18&37548\\
%		\hline
%		chess       &3196&36&115056\\
%		\hline
%		HillValley&1212&100&121200\\
%		\hline
%		HTRU2 &17898&8&143184\\
%		\hline
%	\end{tabular}
%\end{table}

\begin{figure*}[!tb]%\vspace{-2mm}
	\begin{center}
		\begin{minipage}{0.31\linewidth}
			\includegraphics[width=1\textwidth]{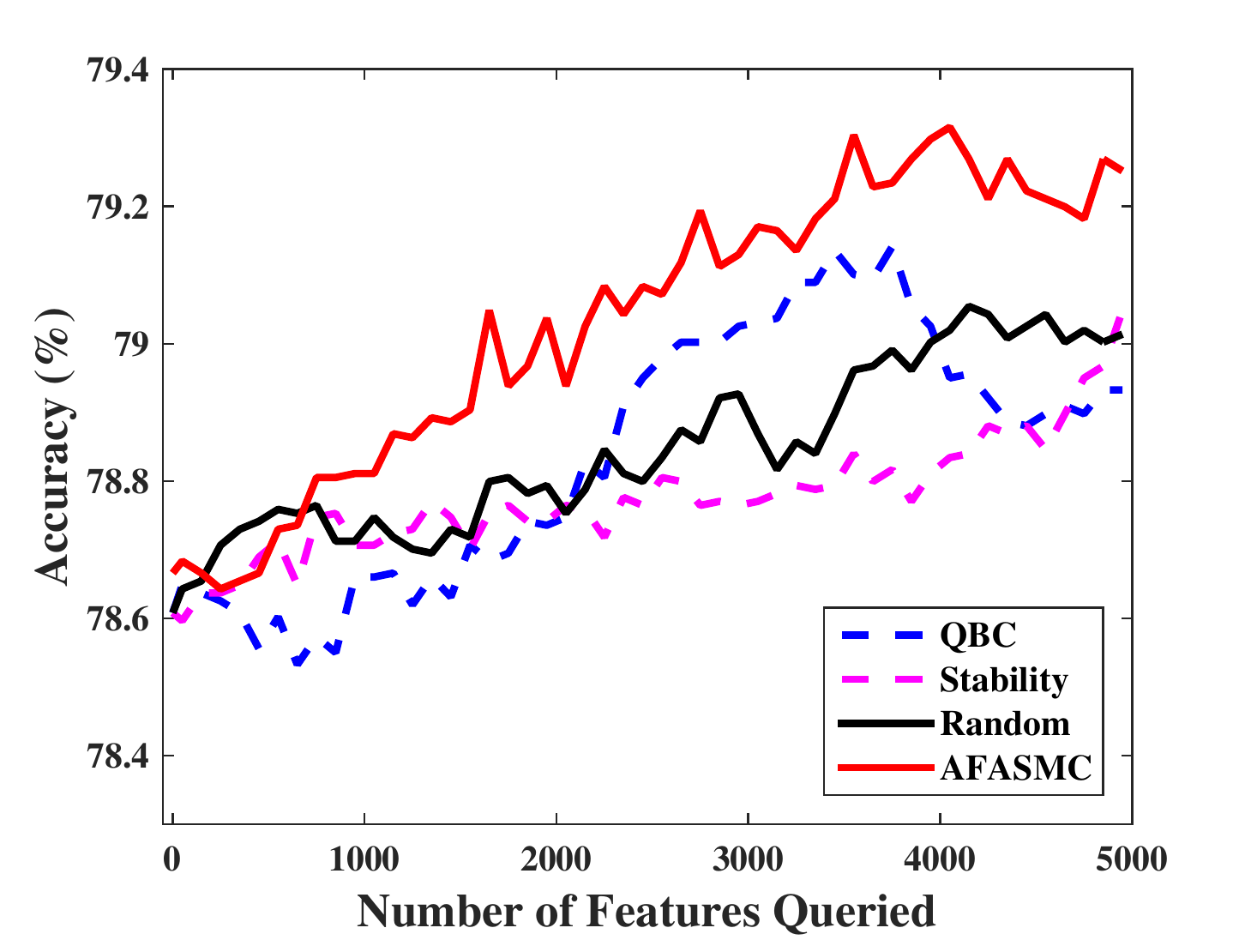}\\
			\centering{(a) abalone}
		\end{minipage}$\quad$
		\begin{minipage}{0.31\linewidth}
			\includegraphics[width=1\textwidth]{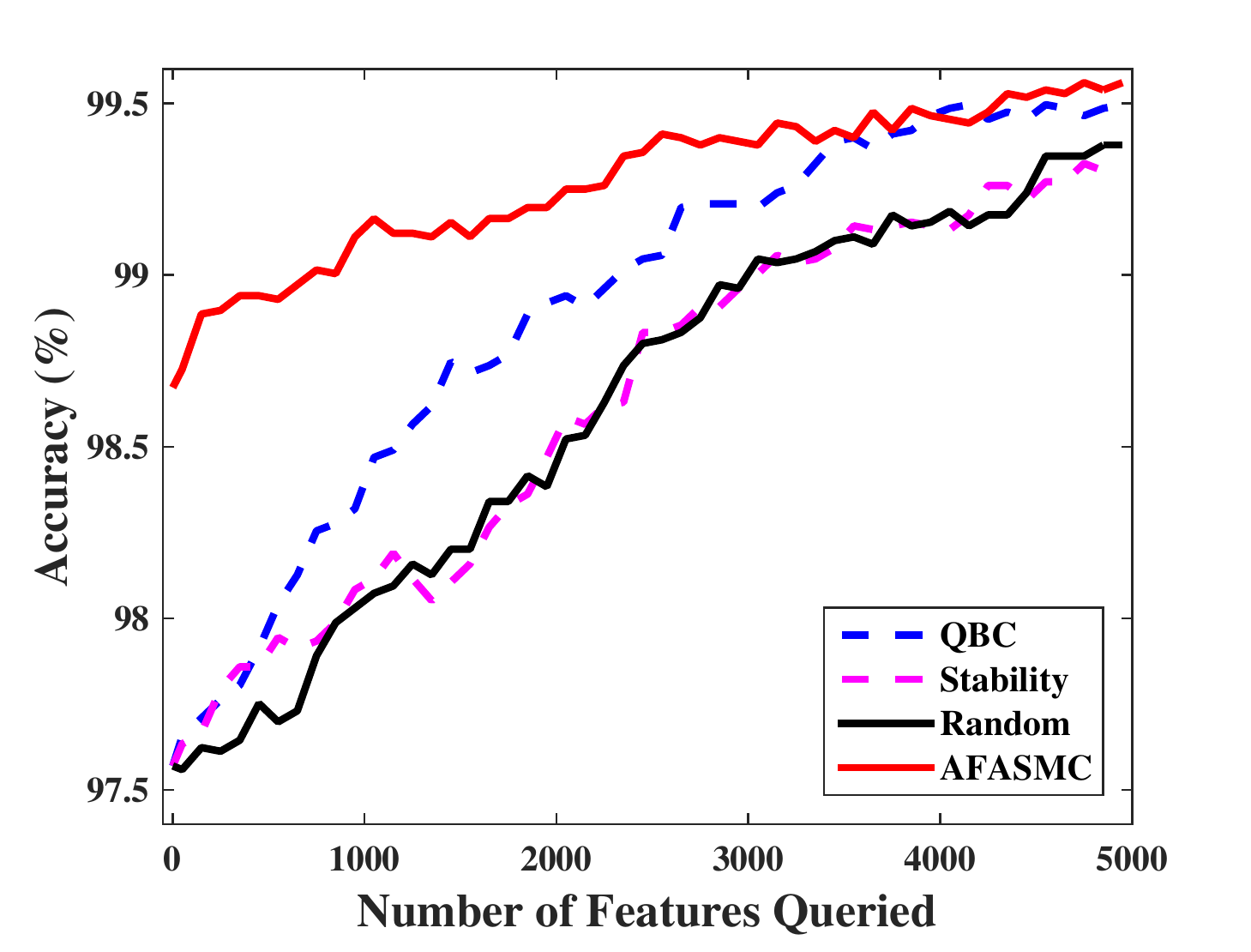}\\
			\centering{(b) letter}
		\end{minipage}$\quad$
		\begin{minipage}{0.31\linewidth}
			\includegraphics[width=1\textwidth]{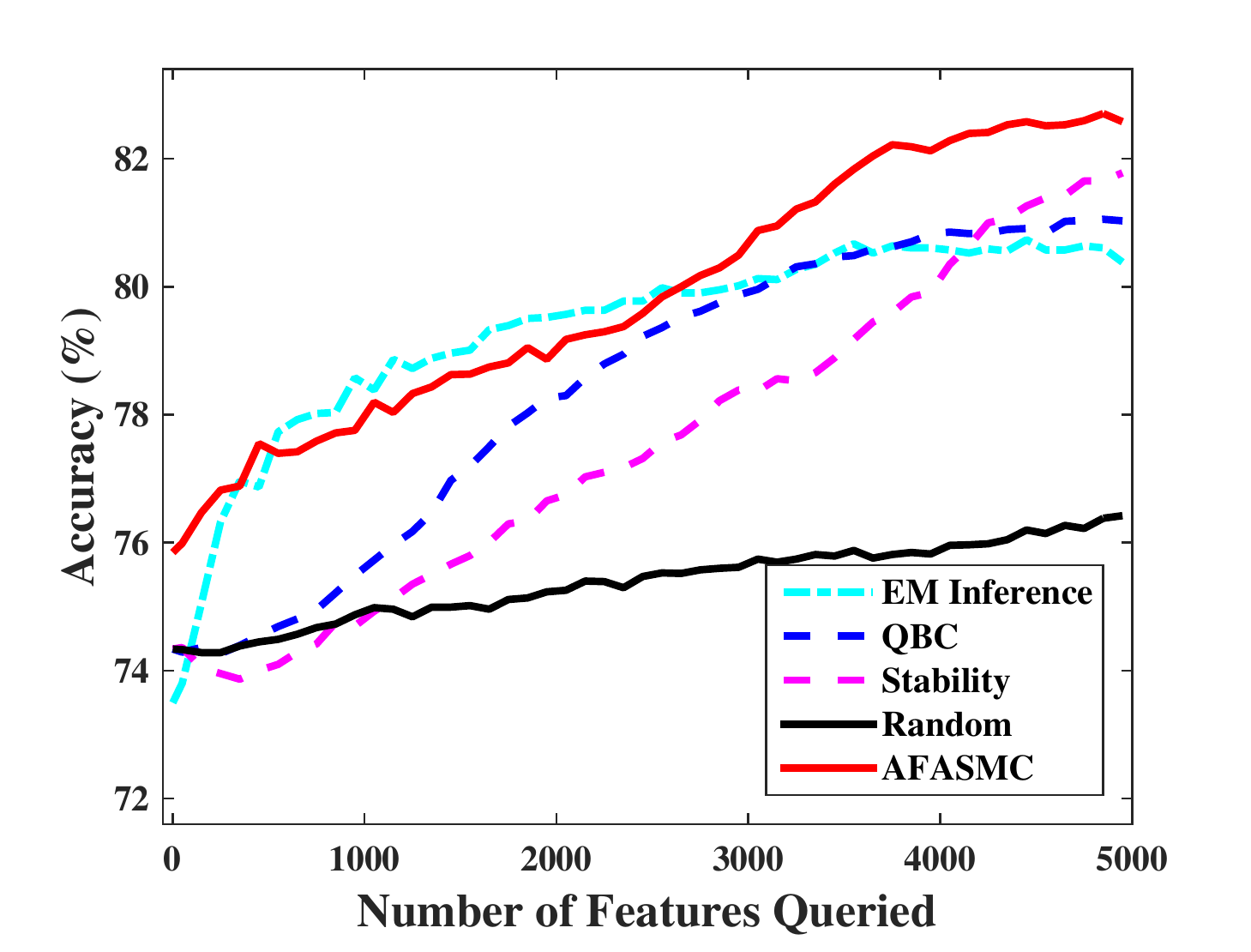}\\
			\centering{(c) image}
		\end{minipage}
		
		\begin{minipage}{0.31\linewidth}
			\includegraphics[width=1\textwidth]{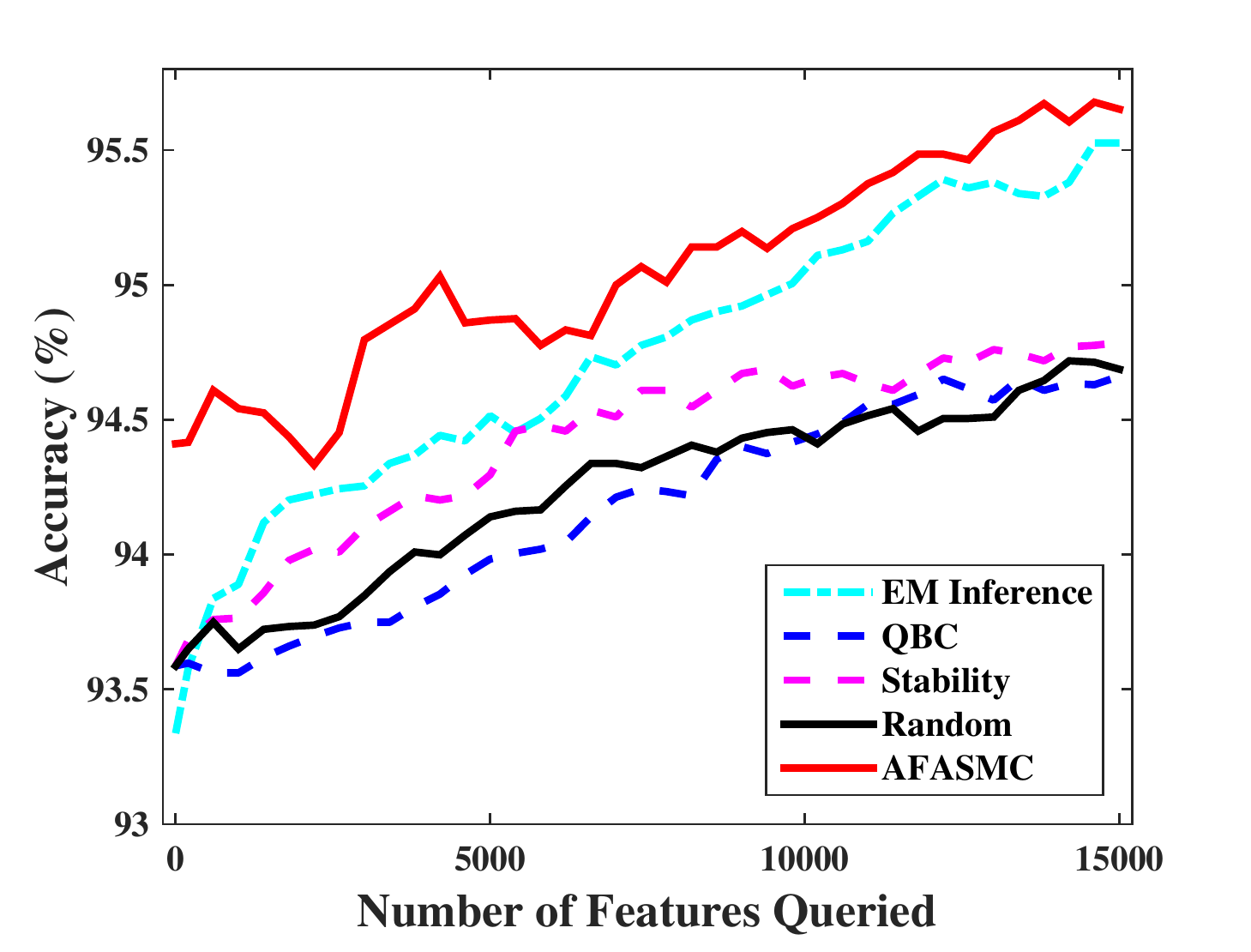}\\
			\centering{(d) chess}
		\end{minipage}$\quad$
		\begin{minipage}{0.31\linewidth}
			\includegraphics[width=1\textwidth]{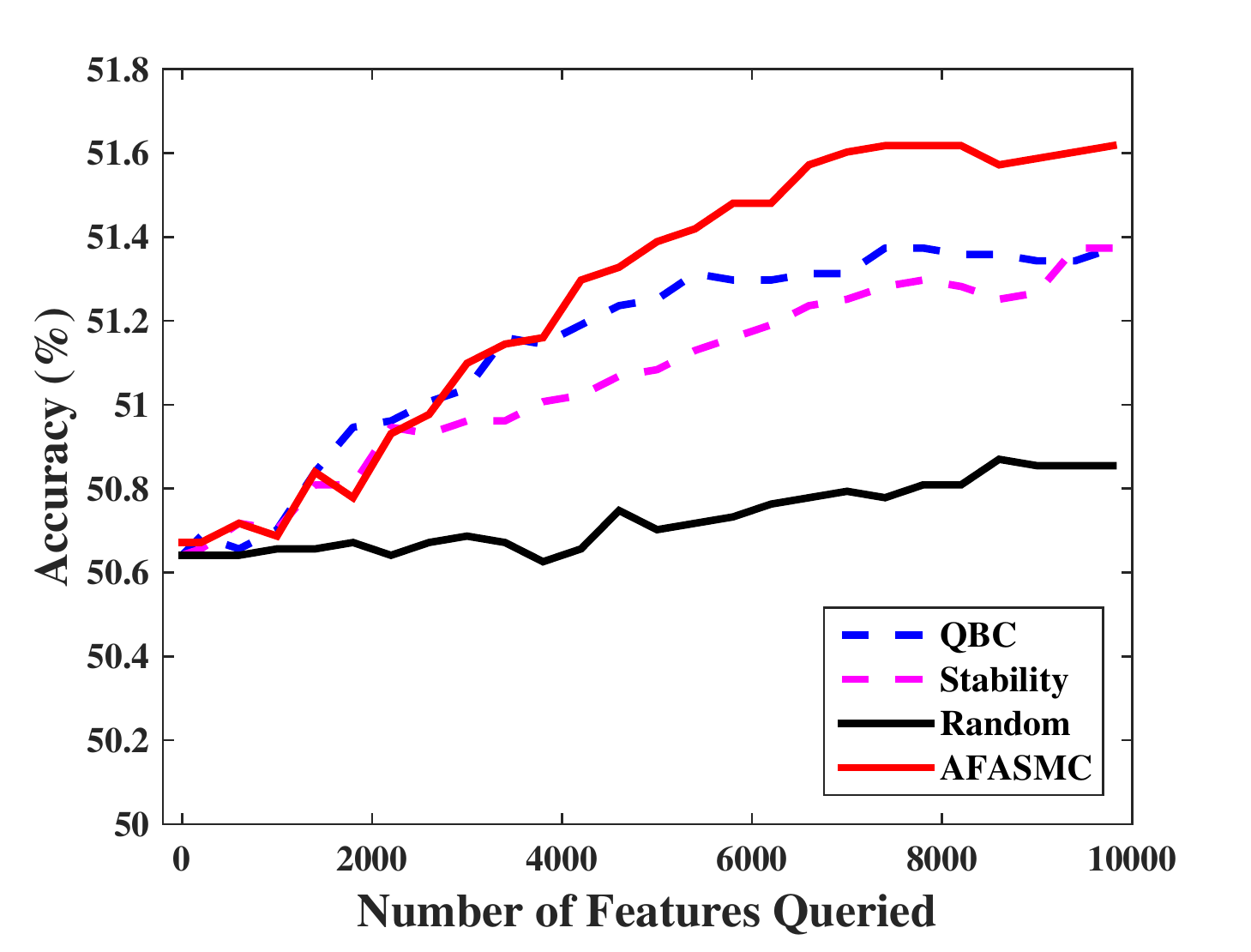}\\
			\centering{(e) HillValley}
		\end{minipage}$\quad$
		\begin{minipage}{0.31\linewidth}
			\includegraphics[width=1\textwidth]{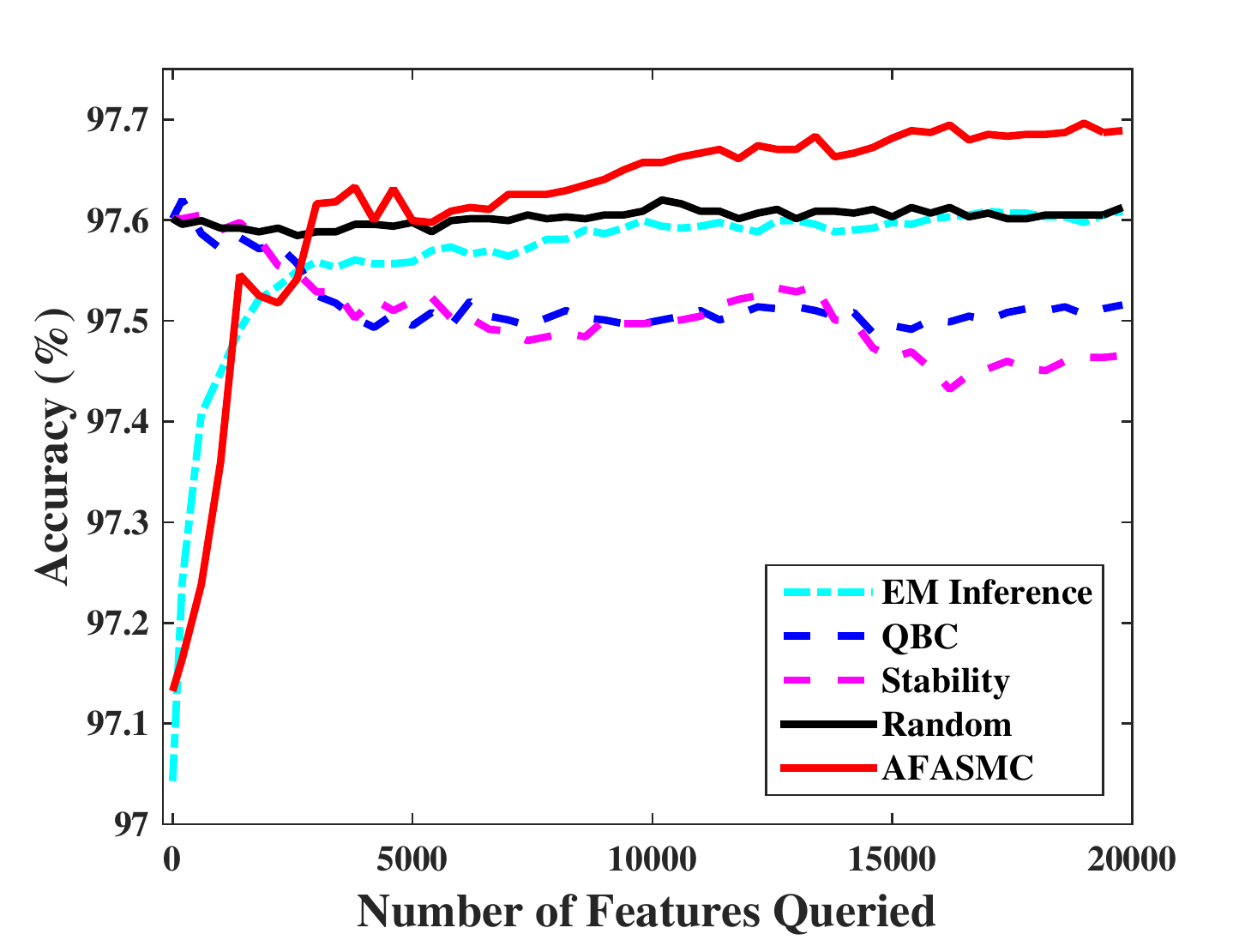}\\
			\centering{(f) HTRU2}
		\end{minipage}
	\end{center}
	\caption{The accuracy curves of compared methods with the number of queried features increasing. }
	\label{fig:active}
\end{figure*}

\begin{table*}[!thb]
	\centering
	\caption{The AUC results. The best performances based on paired $t$-tests at 95\% significance level are bold.} \label{table:auc}
	\renewcommand{\multirowsetup}{\centering}
	\begin{tabular}{c  c  c  c  c  c  c c c }
		\hline
		\multirow{2}{*}{Data}& \multirow{2}{*}{Algorithms} &\multicolumn{7}{c}{Percentage of queried entries}\\
		\cline{3-9}
		&  & 5\%&10\%&20\%&30\%&40\%&50\%&80\%\\
		\hline
		
		\multirow{4}{*}{abalone} &Random &0.859$\pm$0.010 &0.859$\pm$0.010 &0.860$\pm$0.010 &0.861$\pm$0.009 &0.861$\pm$0.009 &0.862$\pm$0.009 &0.866$\pm$0.009\\ 
		&QBC &0.858$\pm$0.010 &0.858$\pm$0.010 &0.859$\pm$0.009 &0.859$\pm$0.009 &0.861$\pm$0.009 &0.864$\pm$0.009 &\textbf{0.868$\pm$0.009}\\ 
		&Stability &0.859$\pm$0.010 &0.859$\pm$0.009 &0.859$\pm$0.009 &0.860$\pm$0.009 &0.861$\pm$0.008 &0.862$\pm$0.008 &0.865$\pm$0.009\\ 
		&AFASMC &\textbf{0.862$\pm$0.009}&\textbf{0.862$\pm$0.009}&\textbf{0.863$\pm$0.009}&\textbf{0.865$\pm$0.009}&\textbf{0.866$\pm$0.010}&\textbf{0.867$\pm$0.010}&\textbf{0.867$\pm$0.010}\\ 
		\hline
		\multirow{4}{*}{letter} &Random &0.999$\pm$0.001 &0.999$\pm$0.000 &0.999$\pm$0.000 &1.000$\pm$0.000 &1.000$\pm$0.000 &1.000$\pm$0.000 &\textbf{1.000$\pm$0.000}\\ 
		&QBC &\textbf{0.999$\pm$0.000}&\textbf{1.000$\pm$0.000}&\textbf{1.000$\pm$0.000}&\textbf{1.000$\pm$0.000}&\textbf{1.000$\pm$0.000}&\textbf{1.000$\pm$0.000}&\textbf{1.000$\pm$0.000}\\ 
		&Stability &0.999$\pm$0.001 &0.999$\pm$0.001 &1.000$\pm$0.000 &\textbf{1.000$\pm$0.000}&\textbf{1.000$\pm$0.000}&\textbf{1.000$\pm$0.000}&\textbf{1.000$\pm$0.000}\\ 
		&AFASMC &\textbf{1.000$\pm$0.000}&\textbf{1.000$\pm$0.000}&\textbf{1.000$\pm$0.000}&\textbf{1.000$\pm$0.000}&\textbf{1.000$\pm$0.000}&\textbf{1.000$\pm$0.000}&\textbf{1.000$\pm$0.000}\\ 
		\hline
		\multirow{4}{*}{image} &Random &0.806$\pm$0.012 &0.808$\pm$0.012 &0.810$\pm$0.013 &0.814$\pm$0.014 &0.819$\pm$0.012 &0.824$\pm$0.013 &0.846$\pm$0.013\\ 
		&QBC &0.804$\pm$0.015 &0.811$\pm$0.014 &0.826$\pm$0.012 &0.838$\pm$0.015 &0.845$\pm$0.015 &0.851$\pm$0.013 &\textbf{0.858$\pm$0.012}\\ 
		&Stability &0.805$\pm$0.013 &0.807$\pm$0.014 &0.818$\pm$0.015 &0.829$\pm$0.016 &0.835$\pm$0.015 &0.846$\pm$0.016 &0.857$\pm$0.011\\ 
		&AFASMC &\textbf{0.822$\pm$0.017}&\textbf{0.832$\pm$0.015}&\textbf{0.846$\pm$0.012}&\textbf{0.852$\pm$0.013}&\textbf{0.855$\pm$0.012}&\textbf{0.857$\pm$0.012}&\textbf{0.857$\pm$0.012}\\ 
		\hline
		\multirow{4}{*}{chess} &Random &0.983$\pm$0.004 &0.984$\pm$0.004 &0.986$\pm$0.003 &0.987$\pm$0.003 &0.988$\pm$0.002 &0.989$\pm$0.002 &0.991$\pm$0.002\\ 
		&QBC &0.983$\pm$0.004 &0.984$\pm$0.003 &0.986$\pm$0.004 &0.988$\pm$0.003 &\textbf{0.990$\pm$0.002}&\textbf{0.991$\pm$0.002}&\textbf{0.992$\pm$0.002}\\ 
		&Stability &0.985$\pm$0.004 &\textbf{0.986$\pm$0.004}&0.987$\pm$0.003 &0.987$\pm$0.003 &0.988$\pm$0.002 &0.990$\pm$0.002 &\textbf{0.991$\pm$0.002}\\ 
		&AFASMC &\textbf{0.987$\pm$0.004}&\textbf{0.988$\pm$0.003}&\textbf{0.989$\pm$0.004}&\textbf{0.991$\pm$0.003}&\textbf{0.991$\pm$0.003}&\textbf{0.992$\pm$0.002}&\textbf{0.992$\pm$0.002}\\ 
		\hline
		\multirow{4}{*}{HillValley} &Random &\textbf{0.452$\pm$0.094}&\textbf{0.451$\pm$0.094}&\textbf{0.449$\pm$0.093}&\textbf{0.446$\pm$0.094}&\textbf{0.445$\pm$0.090}&0.442$\pm$0.090&0.437$\pm$0.078\\ 
		&QBC &0.449$\pm$0.078 &\textbf{0.443$\pm$0.075}&\textbf{0.436$\pm$0.075}&0.435$\pm$0.073 &0.434$\pm$0.072 &0.434$\pm$0.072 &0.434$\pm$0.072\\ 
		&Stability &0.445$\pm$0.085 &0.442$\pm$0.086 &0.439$\pm$0.085 &0.438$\pm$0.070 &0.434$\pm$0.072 &0.434$\pm$0.072 &0.434$\pm$0.072\\ 
		&AFASMC &\textbf{0.454$\pm$0.082}&\textbf{0.447$\pm$0.085}&\textbf{0.446$\pm$0.073}&\textbf{0.451$\pm$0.077}&\textbf{0.451$\pm$0.078}&\textbf{0.450$\pm$0.078}&\textbf{0.465$\pm$0.042}\\ 
		\hline
		\multirow{4}{*}{HTRU2} &Random &0.971$\pm$0.002 &\textbf{0.971$\pm$0.002}&0.972$\pm$0.002 &0.972$\pm$0.002 &0.972$\pm$0.002 &0.972$\pm$0.002 &0.973$\pm$0.002\\ 
		&QBC &0.971$\pm$0.002 &0.968$\pm$0.003 &0.969$\pm$0.002 &0.974$\pm$0.002 &0.975$\pm$0.002 &\textbf{0.975$\pm$0.002}&\textbf{0.976$\pm$0.002}\\ 
		&Stability &0.970$\pm$0.002 &0.971$\pm$0.002 &0.969$\pm$0.002 &0.969$\pm$0.003 &0.971$\pm$0.003 &0.971$\pm$0.003 &0.975$\pm$0.002\\ 
		&AFASMC &\textbf{0.972$\pm$0.002}&\textbf{0.971$\pm$0.002}&\textbf{0.973$\pm$0.002}&\textbf{0.975$\pm$0.002}&\textbf{0.975$\pm$0.002}&\textbf{0.976$\pm$0.002}&\textbf{0.976$\pm$0.002}\\ 
		\hline
	\end{tabular}
\end{table*}

In the experiments, we examine the performance both on matrix completion and the classification after active queries. The proposed supervised matrix completion algorithm AFASMC is compared with following methods: OptSpace \cite{KR10}---a low-rank matrix completion method based on spectral techniques and manifold optimization; LmaFit \cite{WZ12}---a low-rank factorization model based on the nonlinear successive over-relaxation (SOR) algorithm; NNLS \cite{TK10}---an accelerated proximal gradient algorithm for low-rank matrix completion.

Also our active feature acquisition method AFASMC is compared with the following methods: QBC \cite{CS13}---an active matrix completion using Query by Committee strategy; Stability \cite{CS13}---an active matrix completion method based on committee stability; EM Inference \cite{MS14}---it selects the instances with maximum expected utility; Random---randomly select features.

For AFASMC, the parameters $\lambda_1$ and $\lambda_2$ are fixed to 1 as default on all datasets. For other methods, parameters are set or tuned as suggested in the corresponding literature. We employ the linear SVM with default parameters as the classifier for all baselines.

\subsection{Results on matrix completion}
Firstly, we examine the effectiveness of  the proposed method for supervised matrix completion. The performances are evaluated with the matrix reconstruct error as well as the classification accuracy. For each dataset, we compare all the methods under different missing rates. The results are reported in Table \ref{table:mc}. The first row of each dataset corresponds to the case where 60\% entries of the training set are observed, while the second row corresponds to the case with 80\% entries observed. From the table we can see that our proposed method AFASMC can achieve the best performance in terms of both reconstruction and classification. The only exception is on HillValley with 60\% observed entries, where AFASMC is outperformed by NNLS on the reconstruction error with tiny margin, but still achieves the best performance on the test accuracy.

\subsection{Results on classification performance}
In this subsection we examine the performance of active feature acquisition. The feature matrix is initialized with 60\% observed entries for each dataset, while the 40\% entries are randomly missing. Then active selection is performed iteratively based on the variance criterion. After each query, we perform matrix completion, and then train a linear SVM on the training data. The accuracy of the classifier on the test set is record.

Figure \ref{fig:active} plots the performance curves of compared methods as the number of queried features increases. Note that the performance of EM Inference is unbearably poor on the abalone, letter and HillValley datasets, and its curves are not plotted on these three datasets to avoid the poor visualization of other curves. Also it can be seen that the initial points are different because the methods are employing different matrix completion methods. It can be observed that the proposed approach AFASMC achieves the best performance in most cases. The performance of EM inference is not stable. It achieves decent performance on image and chess, but loses its edge on the others. The QBC and Stability methods perform similarly and are less competitive to AFASMC in most cases. Lastly, as expected, the Random method is not effective compared to the active methods. We also present in Table \ref{table:auc} the AUC results after different percentages of entires queried. It can be observed that the proposed approach outperforms the others in most cases.

\subsection{Study with varied acquisition costs}
As discussed previously, the acquisition costs of different features may be diverse. In this subsection, we examine the performance of the proposed strategies for cost-effective feature acquisition. We compare the two optional methods: AFASMC+Cost1, which simply divides the informativeness by the cost; and AFASMC+Cost2, which balances the informativeness and cost via bi-objective optimization. We specify the acquisition cost of each feature dimension as a random integer in $\{1, \ldots, 10\}$. Due to space limit, we present the results on the largest dataset HTRU2 as an example.

\begin{figure}
	\begin{center}
		\begin{minipage}{0.8\linewidth}
			\includegraphics[width=1\textwidth]{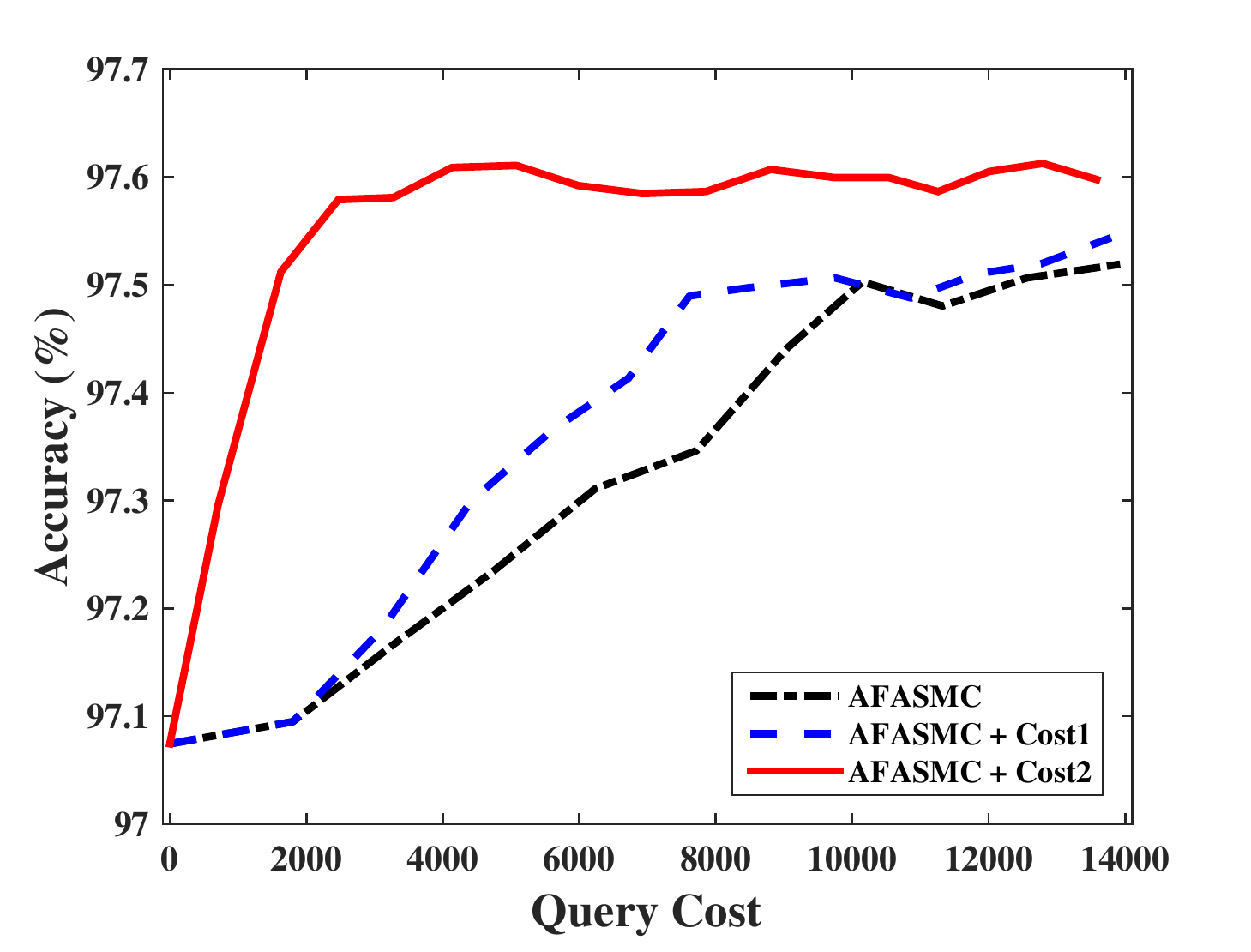}
		\end{minipage}
	\end{center}
	\caption{Results of cost-aware selections on HTRU2.}
	\label{fig:cost}
\end{figure}

We record the accuracy after each query,  and plot the performance curves in Figure \ref{fig:cost}. Note that the curve of the original AFASMC is also presented for reference. It can be observed that both the two strategies for considering the acquisition cost can achieve better performance than the original AFASMC. When comparing AFASMC+Cost1 and AFASMC+Cost2, the method with bi-objective optimization achieves a significantly better performance.

\subsection{Study on the variance computation}
In Section 3.2, when calculating the informativeness based on the variance, we count in all previous iterations of active learning. As discussed before, it is more important to capture the change of an entry within recent iterations. An entry with large variance in the beginning iterations may have been well recovered from recent queries. To examine this idea, we perform experiments to compare the results of calculating the variance with different iterations. Specifically, we use the values of an entry during the last $m$ iterations to calculate the variance, and set $m$ to 2, 4, 8, 16, respectively. Again, for space limit, we report the results on the largest dataset HTRU2 as an example.

\begin{figure}
	\begin{center}
		\begin{minipage}{0.8\linewidth}
			\includegraphics[width=1\textwidth]{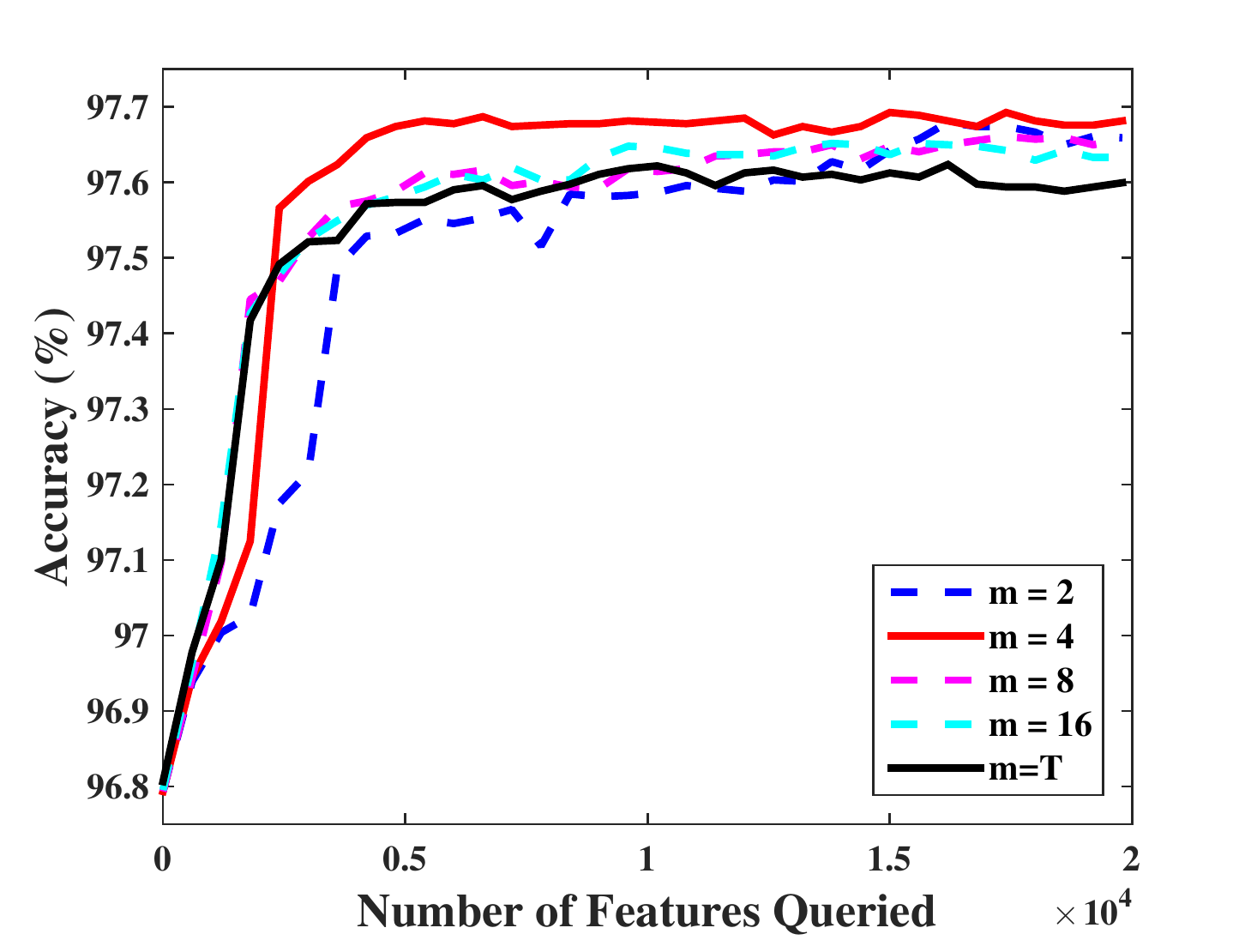}
		\end{minipage}
	\end{center}
	\caption{Comparison of variance computation on HTRU2.}
	\label{fig:variance}
\end{figure}

The performance curves are plotted in Figure \ref{fig:variance}. We also plot the curve of counting all iterations as the original AFASMC method. It can be seen that $m=4$ is the best choice, while counting too few or too many iterations may degrade the performance. This observation is consistent with our conjecture, that the variance computing should emphasize more on recent iterations. Note that we set $m=T$ as default on all datasets to perform the experiments. All the results of AFASMC in previous sections are obtained by counting all iterations. It is thus expected to further improve the performance of the proposed approach by tuning the number of iterations for evaluating the informativeness.

%other possible experiments (optional): To show an image, plot the pixels from dark to shallow in the order of variance.  Low rank assumption validation

\section{Conclusion}
In this paper, we studied the problem of learning from data with missing features. Since the acquisition of ground-truth feature values is usually expensive, our target was to train an effective classification model with the least acquisition cost. We proposed a unified framework to jointly perform matrix completion and active feature acquisition. On one hand, missing values of the feature matrix are recovered by supervised matrix completion, which exploits the feature correlations with a low-rank regularizer, and  the label supervision is utilized by minimizing the empirical classification error. On the other hand, the missing entries are actively queried based on a novel selection criterion, which simultaneously evaluates potential contribution of a feature on both recovering other entries and improving the classification model. Moreover, a bi-objective optimization method was introduced to handle the case where acquisition costs vary for different features. Extensive experimental results validated the superiority of our approach on matrix completion as well as classification performance. In the future, we plan to extend our approach and theoretical analysis to perform active querying both for missing features and class labels.

\section*{Acknowledgment}
This research was partially supported by National Key R\&D Program of China (2018YFB1004300), NSFC (61503182, 61732006), JiangsuSF (BK20150754), the Collaborative Innovation Center of Novel Software Technology and Industrialization, the International Research Center for Neurointelligence (WPI-IRCN) at The University of Tokyo Institutes for Advanced Study. Authors want to thank Bo-Jian Hou for proofreading.

\bibliographystyle{ACM-Reference-Format}
\bibliography{ref}

\end{document}